\documentclass[conference]{IEEEtran}
\usepackage[linesnumbered, ruled]{algorithm2e}
\usepackage{amsmath}
\usepackage{amssymb}
\usepackage{url}
\usepackage{color}
\usepackage{graphicx}
\usepackage{fancyvrb}
\usepackage{float}
\usepackage{subcaption}
\usepackage{balance}
\usepackage{booktabs} 
\usepackage{listings}
\usepackage{wrapfig}
\usepackage{multirow}
\usepackage{footnote}
\usepackage{siunitx}

\usepackage[noadjust]{cite}

\usepackage{siunitx}
\definecolor{mGreen}{rgb}{0,0.6,0}
\definecolor{mGray}{rgb}{0.5,0.5,0.5}
\definecolor{mPurple}{rgb}{0.58,0,0.82}
\definecolor{backgroundColour}{rgb}{0.95,0.95,0.92}

\lstdefinestyle{CStyle}{
    backgroundcolor=\color{backgroundColour},   
    commentstyle=\color{mGreen},
    keywordstyle=\color{magenta},
    numberstyle=\tiny\color{mGray},
    stringstyle=\color{mPurple},
    basicstyle=\ttfamily\footnotesize,
    breakatwhitespace=false,         
    breaklines=true,                 
    captionpos=b,                    
    keepspaces=true,                 
    numbers=left,                    
    numbersep=5pt,                  
    showspaces=false,                
    showstringspaces=false,
    showtabs=false,                  
    tabsize=2,
    language=C
}

\lstdefinestyle{PythonStyle}{
    language=Python,
    backgroundcolor=\color{backgroundColour},   
    commentstyle=\color{mGreen},
    keywordstyle=\color{magenta},
    numberstyle=\tiny\color{mGray},
    stringstyle=\color{mPurple},
    basicstyle=\ttfamily\footnotesize,
    breakatwhitespace=false,         
    breaklines=true,                 
    captionpos=b,                    
    keepspaces=true,                 
    numbers=left,                    
    numbersep=5pt,                  
    showspaces=false,                
    showstringspaces=false,
    showtabs=false,                  
    tabsize=2
}
\newif\ifdraft
\draftfalse  
                                                                            
\ifdraft
  \newcommand{\zhao}[1]{{\textcolor{red}    { ***Zhao:     #1 }}}
  \newcommand{\edit}[1]{{\textcolor{cyan}    {#1}}}
  \newcommand{\outline}[1]{{\textcolor{blue} { ***Outline:  #1 }}}
  \newcommand{\lei}[1]{{\textcolor{red}      { ***Lei:      #1 }}}
  \newcommand{\ian}[1]{{\textcolor{green}    { ***Ian:      #1 }}}
  \newcommand{\note}[1]{ {\textcolor{red}    { \bf          #1 }}}
  \newcommand{\greg}[1]{{\textcolor{magenta} { ***Greg:     #1 }}}
\else
  \newcommand{\zhao}[1]{}
  \newcommand{\edit}[1]{#1}
  \newcommand{\outline}[1]{}
  \newcommand{\lei}[1]{}
  \newcommand{\ian}[1]{}
  \newcommand{\note}[1]{}
  \newcommand{\greg}[1]{}
\fi

\begin{document}
\bstctlcite{IEEEexample:BSTcontrol}

\title{Convolutional Neural Network Training with Distributed K-FAC}

\author{
\IEEEauthorblockN{
  J. Gregory Pauloski\IEEEauthorrefmark{3}, 
  Zhao Zhang\IEEEauthorrefmark{1}, 
  Lei Huang\IEEEauthorrefmark{1}, 
  Weijia Xu\IEEEauthorrefmark{1},
  Ian T. Foster\IEEEauthorrefmark{5}
}
\IEEEauthorblockA{
  \IEEEauthorrefmark{1}Texas Advanced Computing Center\\
  Email: zzhang, huang, xwj@tacc.utexas.edu
} 
\IEEEauthorblockA{
  \IEEEauthorrefmark{3}University of Texas at Austin\\
  Email: jgpauloski@utexas.edu
} 
\IEEEauthorblockA{
  \IEEEauthorrefmark{5}University of Chicago \& Argonne National Laboratory\\
  Email: foster@uchicago.edu
}
}

\maketitle

\begin{abstract}
Training neural networks with many processors can reduce time-to-solution; however, it is challenging to maintain convergence and efficiency at large scales.
The Kronecker-factored Approximate Curvature (K-FAC) was recently proposed as an approximation of the Fisher Information Matrix that can be used in natural gradient optimizers.
We investigate here a scalable K-FAC design and its applicability in convolutional neural network (CNN) training at scale.
We study optimization techniques such as layer-wise distribution strategies, inverse-free second-order gradient evaluation, and dynamic K-FAC update decoupling to reduce training time while preserving convergence.
We use residual neural networks (ResNet) applied to the CIFAR-10 and ImageNet-1k datasets to evaluate the correctness and scalability of our K-FAC gradient preconditioner. 
With ResNet-50 on the ImageNet-1k dataset, our distributed K-FAC implementation converges to the 75.9\% MLPerf baseline in 18--25\% less time than does the classic stochastic gradient descent (SGD) optimizer across scales on a GPU cluster.

\end{abstract}


\section{Introduction}
Deep learning (DL) methods are having transformative impacts on many areas of society, not least in science and engineering where they are enabling exciting new approaches to problems in many disciplines. 
Convolutional neural networks (CNNs) are widely used for classification, regression, object detection, segmentation, and other tasks.
As the powerful memory and communication architectures of high-performance computing (HPC) systems have been shown to support DL applications well, there is growing interest in both the use of HPC for DL~\cite{you2018imagenet, Codreanu2017, Akiba2017, ying2018image, mikami2018massively, Osawa_2019_CVPR} and the use of DL for scientific applications~\cite{lee2019deepdrivemd, Carrasquilla2017, kates2019predicting}.

Leveraging the massive computing resources of supercomputers to dramatically reduce training time is challenging, especially under the constraint of convergence (e.g., validation accuracy)~\cite{mccandlish2018empirical}.
To this end, researchers have examined the scaling properties of first-order methods such as stochastic gradient descent (SGD)~\cite{bottou2018optimization, you2018imagenet, you2019large, mikami2018massively, Akiba2017, ying2018image} and shown promising scaling results for ResNet-50~\cite{he2016deep} and BERT~\cite{devlin2018bert} applications, albeit via the use of application- or architecture-specific techniques such as distributed batch normalization and communication optimization.
A second direction is to explore second-order information such as the Fisher Information Matrix (FIM) to reduce the required number of training iterations. 
Scientists have also examined the natural gradient method (a category of second-order methods) with Kronecker-factored Approximate Curvature (K-FAC)~\cite{martens2015optimizing, Osawa_2019_CVPR, ma2019inefficiency}.
Although previous K-FAC research has shown significant reductions in training iterations and high scalability (up to \num{1024} Nvidia GPUs), the K-FAC implementation used in these works cannot converge to the acceptance performance of MLPerf~\cite{mlperf}, e.g., 75.9\% validation accuracy for ResNet-50, or compete with the faster training times of SGD.

Equations~\ref{eq:sgd} and~\ref{eq:kfac} show the update rules for SGD and the iterative second-order method used in K-FAC, respectively.
$w^{(k)}$ is the weight at iteration $k$, ${\alpha}^{(k)}$ is the learning rate at iteration $k$, $n$ is the mini-batch size, $\nabla{L_i}(w^{(k)})$ is the gradient of the loss function $L_i$ for the $i^\text{th}$ example with regard to $w^{(k)}$, and $F^{-1}(w^{(k)})$ is the inverse of the Fisher information matrix (FIM).

\begin{align}
\text{SGD:  } w^{(k+1)} &= w^{(k)} - \frac{{\alpha}^{(k)}}{n}\sum_{i=1}^{n}\nabla{L_i}(w^{(k)}) \label{eq:sgd}\\
\text{K-FAC: } w^{(k+1)} &= w^{(k)} - \frac{{\alpha}^{(k)}F^{-1}(w^{(k)})}{n}\sum_{i=1}^{n} \nabla{L_i}(w^{(k)}) \label{eq:kfac}
\end{align}

We report in this paper on methods for improving the 
correctness and scalability of K-FAC based optimizers.
We introduce a distributed K-FAC optimization strategy that leverages the independence between layer inputs to make efficient use of a large number of processors.
We study the impact of various explicit and implicit matrix inverse algorithms on training convergence and cost to evaluate $F^{-1}(w^{(k)})\sum_{i=1}^{n}\nabla{L_i}(w^{(k)})$.
Approximating the FIM as a preconditioner to the gradient is expensive; thus, we exploit a conventional method in L-BFGS~\cite{liu1989limited} that decouples the approximation from variable update.
This approach allows us to determine the best preconditioner update frequency to reduce training time while preserving convergence.

Specifically, we design the distribution strategies and study their scaling properties.
The core idea is to distribute per-layer K-FAC calculations to individual processors, then aggregate the results as a preconditioner $F^{-1}$ to the gradient $\nabla{L}(w^{(k)})$ for the iterative second-order method as shown in Equation~\ref{eq:kfac}.
We examine two methods of computing the inverse of $F$, explicit inverse and implicit eigen decomposition, and select the latter algorithm as it preserves training convergence at scale.
We also integrate a set of techniques including dynamic K-FAC update decoupling, damping decay, and K-FAC update frequency decay to reduce training time while preserving model convergence.
Details are presented in \S\ref{sec:design} and \S\ref{sec:impl}.

We prototype our solution in the widely adopted PyTorch DL framework~\cite{NEURIPS2019_9015} and Horovod distributed training framework~\cite{sergeev2018horovod}.
We use ResNet-34, ResNet-50, ResNet-101, and ResNet-152 as example applications and train on various numbers of Nvidia V100 GPUs on the TACC Frontera supercomputer.
We show that with our new method, ResNet-50 on the ImageNet-1k~\cite{Deng2009} dataset converges to above or equal to the 75.9\% baseline performance required by MLPerf~\cite{mlperf} at all scales.
Across scales, our K-FAC based optimization is 18--25\% faster than SGD.

This work makes four contributions:
\begin{itemize}
\item{} A distributed K-FAC optimization strategy;
\item{} A study of explicit/implicit matrix inverse algorithms on its impact to K-FAC and training convergence;
\item{} An empirically optimal preconditioner update frequency for example applications;
\item{} An open source implementation of the proposed algorithm using PyTorch~\cite{NEURIPS2019_9015} and Horovod~\cite{sergeev2018horovod}.
\end{itemize}

The rest of this paper is organized as follows: 
\S\ref{sec:back} discusses parallel DL training, SGD, and K-FAC. 
We review and summarize previous work in scalable DL training in \S\ref{sec:related}. 
\S\ref{sec:design} introduces key system design options and decisions.
\S\ref{sec:impl} discusses technical decisions in details.
\S\ref{sec:expe} presents the experiment results, analysis, and discussion. 
We draw conclusions in \S\ref{sec:conc}.

\section{Background}
\label{sec:back}
Neural network training is usually done through an iterative procedure.
Most existing methods exploit the batch optimization method~\cite{bottou2018optimization},
where a mini-batch of training data is fed to neural network to derive an averaged loss and then used to update variables using corresponding rules.
Equations~\ref{eq:sgd} and~\ref{eq:kfac} show the update rules of SGD and K-FAC.

In this section, we review distributed training strategies and explain in detail the applicability of the data parallel approach in supporting SGD and K-FAC.

\subsection{Data Parallelism}
There are three common strategies to distribute DL training across multiple processors: \emph{data parallel}, \emph{model parallel}, and \emph{hybrid parallel}.
The data parallel approach replicates the model across processors and distributes a mini-batch of training data to these processors in each iteration.
The model parallel approach distributes a large model, usually exceeding the capacity of the host memory of a processor, to multiple processors.
A hybrid approach mixes the two above approaches either by applying different parallel approaches on different layers, or by partitioning processors into groups, within which model parallelism is exploited while data parallelism is used across groups.

In practice, the data parallel approach is dominant in DL training at large scale,
as it achieves better machine utilization compared to the model parallel method.
Numerous effort such as Intel MLSL~\cite{MLSL}, Horovod~\cite{sergeev2018horovod}, TensorFlow~\cite{Abadi2016}, and PyTorch~\cite{NEURIPS2019_9015} have native support for data parallel training via NCCL, Gloo~\cite{incubator2017gloo}, or MPI collective operations~\cite{Thakur2005}.

\subsection{SGD}
\label{sec:back:sgd}
The implementation of SGD via data parallelism falls into two categories: synchronous~\cite{ginsburg2018large} and asynchronous~\cite{recht2011hogwild, zhang2015deep, gossiping-jin2016scale, momentum-mitliagkas2016asynchrony, li2014scaling}, depending on whether all variables are updated in every iteration.
Asynchronous SGD methods have been proven to have a non-linear slowdown compared to synchronous SGD~\cite{alistarh2018convergence}.
Thus we only consider synchronous methods in this paper.

\begin{figure}[t]
\begin{center}
    \includegraphics[width=\columnwidth]{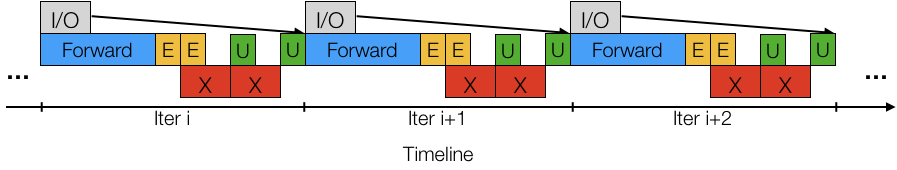}
    \caption{An overview of the synchronous SGD iteration. I/O: Training programs read files. Forward: Forward computation. E: Gradient evaluation. X: Gradient exchange. U: Trainable variable update. \edit{Gradient evaluation and exchange are commonly referred together as back-propagation.}
    \label{fig:streamline}
}
\end{center}
\end{figure}

Synchronous SGD is an iterative procedure of five steps: 1) I/O, 2) forward compute, 3) gradient evaluation, 4) gradient exchange, and 5) variable update.
The distributed training program reads a fixed batch of training items from a storage system and may preprocess them. 
The training items are then fed to the neural network for forward computation to determine the defined loss.
Given the loss, the gradient evaluation step computes the gradient for each trainable variable.
All processors then communicate to exchange the gradients (e.g., to compute the average in SGD).
Finally, the trainable variables are updated with the corresponding gradients using a specific rule, e.g., gradient descent.
An iteration completes when all variables are updated.
Most modern DL frameworks implement this procedure by using a streamlined parallelism with I/O and gradient exchange being performed asynchronously to the other steps, as shown in Figure~\ref{fig:streamline}. 
This approach allows for high utilization of available hardware such as CPUs, GPUs, and interconnects.

With SGD, the only data that is communicated are the initial model weights before training starts and the gradients in each iteration.
Collective operations such as \emph{broadcast()} and \emph{allreduce()} suffice the requirements for SGD.

\subsection{K-FAC}

K-FAC is a method for efficiently approximating the natural gradient. In natural gradient descent optimization, the Fisher information matrix (FIM) is used to represent the curvature of the loss function. While computing the FIM is complex, K-FAC approximates the FIM as Kronecker products of smaller matrices. These smaller matrices are more efficiently invertable. 

The FIM can be interpreted as the negative expected Hessian of a log-likelihood and is given by
\begin{align}
F=\mathbb{E}[\nabla\log p(y|x;\theta)\nabla\log p(y|x;\theta)^T].
\end{align}

\begin{figure}[t]
\begin{center}
    \includegraphics[width=85mm]{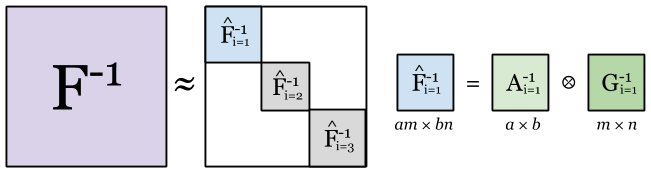}
    \caption{The K-FAC approximation of the Fisher information matrix. $\otimes$ is the Kronecker product.
    \label{fig:kfac}}
\end{center}
\end{figure}

K-FAC approximates $F$ as $\hat{F}$, a diagonal block matrix where each block represents one layer in a network of $L$ layers:
\begin{align}
\hat{F}=\texttt{diag}(\hat{F}_1,...,\hat{F}_i,...,\hat{F}_L)
\end{align}
where
\begin{align}
\hat{F}_i=a_{i-1}a_{i-1}^T\otimes g_ig_i^T=A_{i-1}\otimes G_i.
\end{align}
This is a Kronecker-factorization of $\hat{F}_i$ where the Kronecker factors, often referred to as the covariance matrices, are $A_{i-1}=a_{i-1}a_{i-1}^T$ and $G_i=g_ig_i^T$. 
The activation of the $i-1^\text{th}$ layer and the gradient of the output of the $i^\text{th}$ layer are represented as $a_{i-1}$ and $g_i$ respectively.

The Kronecker product $A \otimes B$ where $A$ has size $m\times n$ and $B$ has size $p \times q$ is:
\begin{align}
A \otimes B=
\begin{bmatrix}
a_{11}B  & \dots  & a_{1n}B \\
\vdots   & \ddots & \vdots  \\
a_{m1}B  & \dots  & a_{mn}B
\end{bmatrix}.
\end{align}
For example:
\begin{align}
\begin{bmatrix}
1 & 2 \\
3 & 4 
\end{bmatrix}
\otimes
\begin{bmatrix}
5 & 6 \\
7 & 8 \\
9 & 0
\end{bmatrix}
=
\begin{bmatrix}
1\times 5 & 1\times 6 & 2\times 5 & 2\times 6 \\
1\times 7 & 1\times 8 & 2\times 7 & 2\times 8 \\
1\times 9 & 1\times 0 & 2\times 9 & 2\times 0 \\
3\times 5 & 3\times 6 & 4\times 5 & 4\times 6 \\
3\times 7 & 3\times 8 & 4\times 7 & 4\times 8 \\
3\times 9 & 3\times 0 & 4\times 9 & 4\times 0
\end{bmatrix}
\end{align}

The inverse of $\hat{F}_i$ can be computed as the Kronecker product of the inverse of the factors $A_{i-1}$ and $G_i$, as shown in Figure \ref{fig:kfac}, using the property $(A\otimes B)^{-1}=A^{-1}\otimes B^{-1}$.
\begin{align}
\hat{F}_i^{-1}=A_{i-1}^{-1}\otimes G_i^{-1}
\end{align}
We can then use $\hat{F}_i^{-1}$ to precondition the gradient, $\nabla L$, of the parameters $w_i^{(k)}$ for layer $i$ and iteration $k$.
\begin{align}
w_i^{(k+1)}=w_i^{(k)}-\alpha^{(k)}\hat{F}_i^{-1}\nabla L_i(w_i^{(k)})
\end{align}
Using the relation $(A\otimes B)\vec{c}=B^T\vec{c}A$, the preconditioned gradient $\hat{F}_i^{-1}\nabla L_i(w_i^{(k)})$ can be efficiently computed as
\begin{align}
\hat{F}_i^{-1}\nabla L_i(w_i^{(k)})=G_i^{-1}\nabla L_i(w_i^{(k)})A_{i-1}^{-1}.
\end{align}

The FIM approximation $\hat{F_i}$ can be ill-conditioned for inverting so to compensate for inherent inaccuracies, a Tikhonov regularization technique is applied where $\gamma I$ is added to $\hat{F}_i$ \cite{martens2016conv, Osawa_2019_CVPR}.
We call $\gamma$ the damping parameter, and $\gamma I$ can be added to the Kronecker factors such that instead of computing $\hat{F}_i^{-1}$, we compute $(\hat{F}_i+\gamma I)^{-1}$ as:
\begin{align}
(\hat{F}_i+\gamma I)^{-1}=(A_{i-1}+\gamma I)^{-1}\otimes (G_i+\gamma I)^{-1}. \label{eq:kfac-damped}
\end{align}
Thus, the final update step for the parameters at iteration $k$ is:
\begin{align}
w_i^{(k+1)}=&w_i^{(k)}-\alpha^{(k)}(G_i+\gamma I)^{-1}\nabla L_i(w_i^{(k)})(A_{i-1}+\gamma I)^{-1}
\end{align}

\subsection{Frameworks}
We choose PyTorch \cite{NEURIPS2019_9015} and Horovod \cite{sergeev2018horovod} to prototype the K-FAC optimizer.
PyTorch is an widely adopted DL framework with comprehensive support of various neural network architectures and high performance.
At its backend, PyTorch uses C++ runtime for performance.
At its frontend, PyTorch exposes a linear algebra interface and a neural network interface, which can be used to implement certain matrix operations and compose neural networks, respectively.
PyTorch has an imperative programming interface which eases our development compared to the symbolic programming interface of TensorFlow \cite{Abadi2016}.
PyTorch has its own distributed training support with collective communication in MPI, NCCL, or Gloo \cite{incubator2017gloo}. 
We choose Horovod \cite{sergeev2018horovod} instead, as Horovod supports PyTorch, TensorFlow, and MXNet~\cite{Chen2015} and we plan to extend our K-FAC optimization to other major DL frameworks.

Horovod is a general communication framework that can call MPI, NCCL, or IBM Distributed Deep Learning Library (DDL) primitives.
It inherits MPI concepts such as \emph{size}, \emph{rank}, and \emph{local rank}, and
exposes a limited communication interface with \emph{allreduce()}, \emph{allgather()}, and \emph{broadcast()} primitives.
In particular, its \emph{allreduce()} operation is implemented by using the scatter-reduce algorithm, which is bandwidth optimal in the ring topology under the assumption that DL gradient exchange involves large enough data to be bandwidth bound \cite{patarasuk2009bandwidth}.
\emph{Allreduce{}} in Horovod can be synchronous or asynchronous; users can specify the size of fusion buffer that accumulates data before communication until reaching the buffer size.
The fusion buffer is usually set as 16~MB or 32~MB to guarantee that each \emph{allreduce()} is bandwidth dominated.
Thus the communication in Horovod can make highly efficient use of the underlying interconnect.

\section{Related Work}
\label{sec:related}
Much recent research has focused on scaling DL training via the use of large batch sizes. For example, with the classic image classification problem, a batch size of 32K is considered large for convolutional neural network training with the ImageNet-1k dataset. 
Large batches make it possible to distribute enough data to a large number of processors to maintain high utilization.
However, they lead to high communication costs at large scale, as the size of the gradients are large enough that the communication overhead is dominated by bandwidth and is proportional to the number of processors.

In this section, we summarize previous work of DL training at large scale and justify the fundamental difference between our work and previous efforts.

\subsection{Scaling results of SGD}
Previous research such as LARS~\cite{you2018imagenet} and LAMB~\cite{you2019large} propose SGD variants with layer-wise adaptive learning rate and learning rate schedules to enable large batch size training without losing model convergence.
The ResNet-50 training with ImageNet-1k dataset is shown to converge to 74.9\% baseline at the time in 20 minutes on 2048 Intel Xeon Platinum 8160 processors~\cite{you2018imagenet}.
Subsequent work~\cite{Akiba2017, ying2018image, mikami2018massively} leverages the same optimizer with dedicated optimizations for the interconnect architecture, e.g., 2D torus, and processor architecture optimizations such as GPU and TPU.
Researchers were able to bring the training time to $\sim$2.2 minutes on 1024 TPUs~\cite{ying2018image}, with a model that converges to 76.3\% validation accuracy, which is above the MLPerf~\cite{mlperf} baseline of 75.9\%.
Some techniques such as mixed precision, interconnect-aware \emph{allreduce()}, and distributed batch normalization are specific to certain hardware and applications.
In contrast, our work focuses on general optimizations that can be applied regardless of processor architecture, interconnect topology, or application.

\subsection{Scaling results of K-FAC}
K-FAC~\cite{martens2015optimizing} has been shown to take fewer steps to converge for image classification~\cite{Osawa_2019_CVPR,Ba2017Distributed} and language processing tasks~\cite{martens2018kronecker}. 
However, each step in K-FAC runs significantly slower than in SGD, as the FIM must be approximated.
\edit{In previous work, an asynchronous distributed K-FAC using a doubly-factored Kronecker approximation was used to achieve a time-per-iteration comparable to standard SGD training on ImageNet-1k \cite{Ba2017Distributed} .
While this work reports a 2$\times$ improvement in overall training times due to faster convergence at the start of training, the implementation only converges to a final 70\% validation accuracy.}
Researchers have also published a distributed implementation of K-FAC that distributes a block-diagonal approximation (with each block associated to one layer) across GPUs~\cite{Osawa_2019_CVPR}.
Even though the work reports that their K-FAC implementation takes $\sim$10 minutes to reach the 74.9\% validation accuracy baseline in just 978 iterations,
there is no training time comparison of K-FAC to SGD. 
In addition, the model convergence is obviously 1$\%$ lower than the MLPerf~\cite{mlperf} baseline, which is considered as an acceptance test for DL researchers and practitioners.
In our work, we prioritize training correctness and make such system design decisions when facing performance tradeoffs.
Further, we compare K-FAC training time to SGD and design strategies to reduce training time without losing model convergence.

\begin{figure*}[ht]
\begin{center}
    \includegraphics[width=7in]{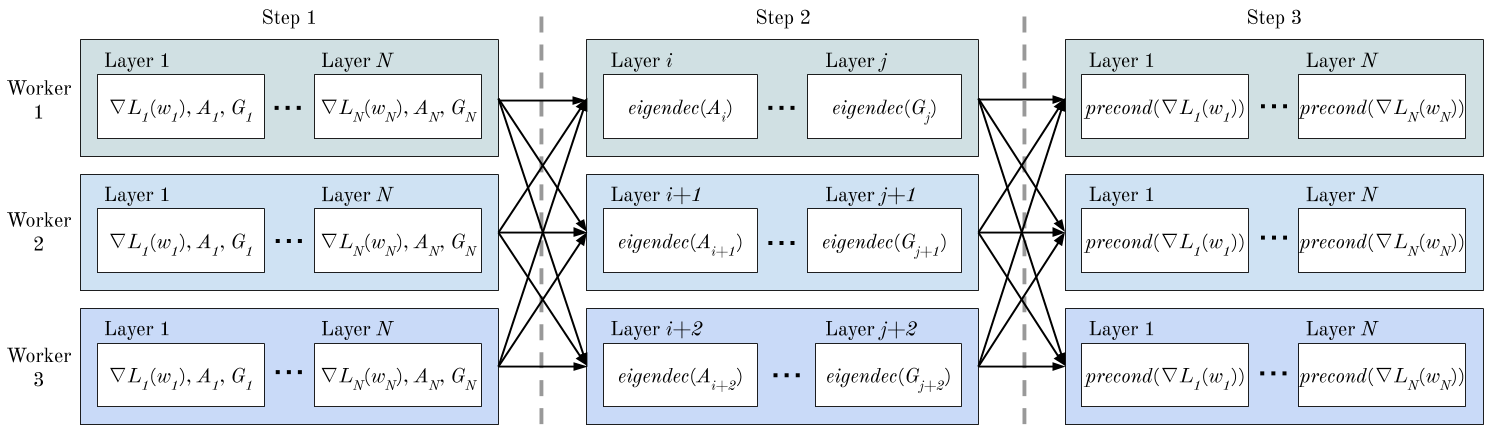}
    \caption{Overview of our layer-wise factor distribution scheme. In step 1, each worker computes the gradient, $\nabla L_i(w_i)$, and the factors, $A_{i-1}$ and $G_{i}$, for each layer using the worker's mini-batch. Before step 2, the gradients and factors are \emph{allreduced}, and each factor is assigned to a worker in a round-robin fashion. In step 2, each worker computes the eigen decomposition of each factor it was assigned. The results from step 2 are gathered across all workers before step 3 where the preconditioned gradient is computed locally and in-place for all layers following Equations \eqref{eq:v1} to \eqref{eq:kfac-eigen}. In step 2, we use indices $i$ for $A$ and $j$ for $G$ to denote that the eigen decomposition for $A_{i-1}$ and $G_{i}$ can occur on different workers. The pseudocode for each step is provided in Algorithm \ref{lst:kfac-algo}.
    \label{fig:communication}}
\end{center}
\end{figure*}

\section{Design}
\label{sec:design}
Recent works have found promising results in reducing the overhead of K-FAC by assigning each worker, e.g., GPU, to compute the FIM approximation for a single layer such that the K-FAC update for each layer can be computed in parallel \cite{Osawa_2019_CVPR}. 
While this method is effective when the number of layers in the network is equal to or greater than the number of workers, the scaling performance decreases as workers are left idle when there are more workers than layers to compute.
Our distributed K-FAC design seeks to improve on this method by reducing the frequency of communication, increase the granularity at which the K-FAC computations can be distributed, and ensuring our design achieves MLPerf baselines on benchmarks. We also design our K-FAC algorithm to act as a gradient preconditioner such that K-FAC can be used in-place with any standard optimizer, such as Adam, LARS, or SGD.

\subsection{Matrix Inversion}
A standard K-FAC update step for one layer requires inverting two matrices, $(A_{i-1}+\gamma I)$ and $(G_i + \gamma I)$, as shown in Equation \eqref{eq:kfac-damped}.
However, \edit{it} has been shown that $(\hat{F}_i+\gamma I)^{-1}$ can be computed implicitly using an alternative method based on the eigendecompostion of $\hat{F_i}$ \cite{martens2016conv}.
Let $A_{i-1}=Q_G{\Lambda}_GQ_G$ and $G_i=Q_A{\Lambda}_AQ_A$ be the eigen decompositions of the factors $A_{i-1}$ and $G_i$.
Then, we can compute the preconditioned gradient as:
\begin{align}
V_1&=Q_G^T L_i(w_i^{(k)}) Q_A \label{eq:v1}\\
V_2&=V_1/(\upsilon_G\upsilon_A^T+\lambda) \label{eq:v2}\\
(\hat{F}_i+\gamma I)^{-1}\nabla L_i(w_i^{(k)})&=Q_GV_2Q_A^T \label{eq:kfac-eigen}
\end{align}
where $\upsilon_A$ and $\upsilon_G$ are vectors of the eigenvalues of $A_{i-1}$ and $G_i$, i.e. the diagonals of $\Lambda_A$ and $\Lambda_G$. 
For the full proof of this eigen decomposition expansion, see Appendix A.2 in \cite{martens2016conv}. 

In our design, we use the eigen decomposition expansion of $(\hat{F}_i+\gamma I)^{-1}$ in Equations \eqref{eq:v1}--\eqref{eq:kfac-eigen} to compute the preconditioned gradient.
In Table \ref{tb:inv_vs_eigen}, we compare the final validation accuracy on CIFAR-10 with ResNet-32 between the explicit inverse method in Equation~\eqref{eq:kfac-damped} and the implicit eigen decomposition method.
We consider the baseline for acceptable accuracy to be 92.49\% \cite{he2016deep}.
As the batch size increases, the inverse K-FAC update performs worse on the validation data and drops below the acceptable baseline whereas both SGD and K-FAC with eigen decomposition perform above baseline performance at all batch sizes.

\begin{table}[]
\begin{center}
    \caption{CIFAR-10 ResNet-32 validation accuracy for inverse vs.\ eigen decomposition K-FAC updates}
    \begin{tabular}{ | c | c | c | c |}
    \hline
    Batch Size            & 256     & 512     & 1024    \\ \hline \hline
    SGD                   & 92.77\% & 92.58\% & 92.69\% \\ \hline
    K-FAC w/ Inverse      & 92.58\% & 92.36\% & 91.71\% \\ \hline
    K-FAC w/ Eigen-decomp. & 92.76\% & 92.90\% & 92.92\% \\ \hline
    \end{tabular}
    \label{tb:inv_vs_eigen}
\end{center}   
\end{table}

\begin{algorithm}
	\DontPrintSemicolon
	\SetArgSty{textnormal}
	\SetDataSty{textnormal}
	\tcc{Compute Gradients}
	\ForEach{worker} {
	  Compute forward and backward pass
	}
	\BlankLine

    \tcc{Step 1: Compute Factors}
	\emph{allreduce}($\nabla L_{1:L}(w_{1:L})$)\; 
	\ForEach{worker} {
	  Compute $A_{0:L-1}$ and $G_{1:L}$
	}
	\emph{allreduce}($A_{0:L-1}$,$G_{1:L}$) \;
	Assign factors $A_{0:L-1}$ and $G_{1:L}$ to unique workers \;
	\BlankLine
	
	\tcc{Step 2: Compute Decompositions}
	\ForEach{worker $w$} { 
	  \ForEach{$A_{i}$ assigned to $w$} { 
	    $Q_{A_i}$, $\Lambda_{A_i}=\emph{eigendecompose}(A_{i})$
	  }
	  \ForEach{$G_{j}$ assigned to $w$} {
	    $Q_{G_j}$, $\Lambda_{G_j}=\emph{eigendecompose}(G_{j})$
	  }
	}
	\emph{allgather}$(Q_{A_{0:L-1}}$, $\Lambda_{A_{0:L-1}}$, $Q_{G_{1:L}}$, $\Lambda_{G_{1:L}})$ \;
	\BlankLine
	
	\tcc{Step 3: Precondition Gradient}
	\ForEach{worker} { 
	  $\emph{precondition}(\nabla L_{1:L}(w_{1:L}))$
	}
	\BlankLine
	
	\tcc{Update Weights with SGD}
	\ForEach{worker} {
	  Update weights using the preconditioned gradients
	}
	\caption{Distributed K-FAC preconditioner}
	\label{lst:kfac-algo}
\end{algorithm}

\subsection{Parallelism}
During each iteration, the forward and backward passes are computed on each worker using the worker's local mini-batch.
Hooks are registered to the input and output of each layer to save the activation of the previous layer and gradient with respect to the output of the current layer.
Then using Horovod's \emph{allreduce()}, the gradients computed in the backward pass are averaged across all workers.

Each worker computes the Kronecker factors $A_{i-1}$ and $G_i$ using the saved activations and gradients of the output for each layer, and we maintain a running average of the Kronecker factors computed from each mini-batch.
\begin{align}
A_{i-1}^{(k)} &= \xi A_{i-1}^{(k)} + (1-\xi) A_{i-1}^{(k-1)} \\
G_i^{(k)}     &= \xi G_i^{(k)}     + (1-\xi) G_i^{(k-1)}
\end{align}
$\xi$ is the running average hyper-parameter typically in the range $[0.9,1)$.
Using \emph{allreduce()}, the updated running averages of the factors are averaged across workers. 
This process of calculating the gradients and factors locally and averaging the results is shown in step 1 of Algorithm \ref{lst:kfac-algo} and Figure \ref{fig:communication}.

Current distributed K-FAC implementations ~\cite{Osawa_2019_CVPR} choose to assign each worker to compute $A_{i-1}^{-1}$, $G_i^{-1}$, and the final preconditioned gradient $(\hat{F}_i+\gamma I)^{-1}\nabla L_i(w_i^{(k)})$ before communicating the per layer preconditioned gradients to all workers such that each worker can update its local weights.
Our design takes a different approach where we assign each worker a single factor to eigen decompose in a round-robin fashion. This is shown in step 2 of Algorithm \ref{lst:kfac-algo} and Figure \ref{fig:communication} where each worker computes the eigen decomposition of the subset of factors $A_{i-1}$ and $G_i$ assigned to itself. 
This approach decouples the eigen decomposition updates from the preconditioning of the gradients.

The eigen decompositions for the factors are then communicated between all workers so that each worker can compute the preconditioned gradients locally using Equations \eqref{eq:v1}--\eqref{eq:kfac-eigen}.
The preconditioned gradients are computed in place so that any standard optimizer, e.g., SGD, can be used to update the weights.

\subsection{Communication}
Our design introduces communication in three places: 1) averaging the gradients, 2) averaging the Kronecker factors, and 3) gathering the eigen decompositions for each factor.
By partitioning the communication into these three parts, we can take advantage of the fact that preconditioned gradients are computed locally to reduce communication in iterations where the factors are not updated.

A common strategy to improve training times when using K-FAC is to only update the Kronecker factors every $n$ iterations.
In iterations where the Kronecker factors are not updated, we can skip the communication for  (2) and (3) and use the stale Kronecker factors from previous iterations that are already stored locally on each worker to compute the preconditioned gradients.
This reduction in communication is possible because we decouple the eigen decomposition of the factors from the preconditioning of the gradients.
In Algorithm \ref{lst:kfac-algo}, this would result in skipping lines 5--18.
As training progresses, the FIM becomes more stable and impact of using stale eigen decompositions decreases.

In practice, the factors can be updated every tens or hundreds of iterations without loss in performance.
Since the K-FAC updates happen infrequently, the majority of training iterations require no extra communication compared to conventional model parallel training with SGD which only requires communicating the gradients.

By decoupling the Kronecker factor calculations from the preconditioned gradient calculation, we are also able to compute the eigen decompositions for $A_{i-1}$ and $G_i$ on different workers and achieve double the worker utilization compared to existing distributed K-FAC implementations that use a layer-wise distribution scheme \cite{Osawa_2019_CVPR}.

\section{Implementation}
\label{sec:impl}
Our implementation is based on an existing open-source K-FAC optimizer written with PyTorch that has no support for distributed training~\cite{wang2019eigen, george2018ekfac}.
We modified this implementation to support our distributed K-FAC design and to act as a preconditioner for standard PyTorch optimizers.
Our implementation supports K-FAC updates for Linear and Conv2D layers.
All unsupported layers are ignored by the K-FAC preconditioner and updated normally using the user's choice of optimizer, such as SGD.
Careful consideration was made to ensure that our K-FAC implementation could be used with minimal changes to existing PyTorch scripts that use Horovod for distributed training.

\subsection{Communication with Horovod}
Existing open-source K-FAC implementations that support distributed training use TensorFlow's parameter server model \cite{KFACTF}.
Parameter servers introduce a bottleneck that inhibits performance at large scale.
For this reason, we use Horovod for communication because Horovod uses a decentralized approach that scales well to thousands of compute nodes ~\cite{sergeev2018horovod}.

Horovod's PyTorch implementation offers support for synchronous and asynchronous communication operations.
All communications operations we used are done asynchronously to take advantage of parallelism between computation and communication. 
Using Horovod, handles are registered to communication operations such that we can register \emph{allreduce()} operations as we compute factors and eigen decompositions across workers and wait to do the communication in batches.

\subsection{K-FAC Interface}
\begin{lstlisting}[style=PythonStyle, label={lst:kfac-code}, caption={Example K-FAC usage.}, float, floatplacement=H]
...

optimizer = optim.SGD(model.parameters(), ...)
optimizer = hvd.DistributedOptimizer(optimizer, ...)
preconditioner = KFAC(model, ...)

...

for i, (data, target) in enumerate(train_loader):
  optimizer.zero_grad()
  output = model(data)
  loss = criterion(output, target)
  loss.backward()

  optimizer.synchronize()
  preconditioner.step()
  with optimizer.skip_synchronize():
    optimizer.step()
    
  ...
\end{lstlisting}
Our K-FAC implementation is designed to be easily inserted into existing training scripts using Horovod.
An example of how to incorporate K-FAC into existing training scripts is given in Listing \ref{lst:kfac-code}.
The only necessary changes are to initialize the \emph{KFAC()} preconditioner and add a call to \emph{KFAC.step()} before \emph{optimizer.step()} as seen in lines 5 and 16 respectively.
By default when using Horovod's \emph{DistributedOptimizer()}, Horovod waits to call \emph{allreduce()} on the gradients until \emph{optimizer.step()} is called. 
However, the gradients must be averaged across workers before we can call \emph{KFAC.step()}, so we call \emph{optimizer.synchronize()}, shown on line 15, before performing the K-FAC preconditioning.

\subsection{K-FAC Hyper-Parameters}
Our K-FAC preconditioner introduces a number of hyper-parameters for controlling gradient clipping, factor update frequency, and damping.

After preconditioning the gradients, we scale the gradients by a factor $\nu$ where
\begin{align}
\nu=\texttt{min}\left( 1,\;
  \sqrt[]{
  \frac{\kappa}
       {\alpha^2\sum_{i=1}^{n}|\mathcal{G}_i^T \nabla L_i(w_i)|}}
\right)
\end{align}
where $\kappa$ is a user-defined constant, typically on the order of $10^{-3}$, $\alpha$ is the learning rate, and $\mathcal{G}_i$ is the preconditioned gradient~\cite{wang2019eigen, george2018ekfac}. 
This gradient scaling is done to prevent the norm of $\mathcal{G}_i$ becoming large compared to $w_i$ \cite{Osawa_2019_CVPR}.

Similar to the work \cite{Osawa_2019_CVPR}, we also use a damping decay scheme whereby we reduce the damping by a fixed scalar quantity at fixed epochs. Starting with a larger damping accounts for rapid changes in the FIM at the start of training.

As training progresses and the FIM becomes more stable, the frequency at which the Kronecker factors and eigen decompositions needs to be updated decreases. We use the hyper-parameter \emph{kfac-update-freq} to control the frequency at which we update the eigen decompositions of $A_{i-1}$ and $G_i$. 
We find that the factors $A_{i-1}$ and $G_{i}$ can be updated and communicated at a frequency of $10 \times $ \emph{kfac-update-freq} without loss in performance.

At fixed training epochs, we decrease \emph{kfac-update-freq } by a scalar quantity to reduce the computation and communication required while preserving accuracy. In practice, we found that it was sufficient to maintain a constant \emph{kfac-update-freq} for the entirety of training, however, small performance improvements can be gained by fine-tuning the update frequency schedule.

\section{Experiments}
\label{sec:expe}
Now we present the empirical results of the scalable K-FAC preconditioner.
In this section, we will introduce the hardware and software platforms, applications, and datasets used in our experiments.
We adopt the MLPerf~\cite{mlperf} acceptance performance as baseline.
We study the correctness, performance, and scalability of our K-FAC preconditioner along with comparisons to SGD.

\subsection{Platforms}
We use the GPU subsystem of the Frontera supercomputer hosted at Texas Advanced Computing Center (TACC).
This machine is powered by IBM Power9 processors and has 112 nodes in total.
There are 4 Nvidia V100 GPUs, 256~GB RAM, and 1~TB rotational disk per node.
The nodes are connected by an InfiniBand EDR network.

We prototype the K-FAC preconditioner using PyTorch v1.1 and Horovod v0.19.0. 
These software frameworks rely on CUDA 10.0, CUDNN 7.6.4, and NCCL 2.4.7.
\edit{We use single-precision floating point numbers (FP32) for training and communication.} Our code is open source with the MIT license and is hosted at \url{https://github.com/gpauloski/kfac_pytorch}.

\subsection{Datasets and Applications}
Throughout the development process, we use ResNet-34~\cite{he2016deep} with the CIFAR-10~\cite{krizhevsky2009learning} dataset to test correctness. 
Then we use the ImageNet-1k dataset~\cite{Deng2009} with ResNet-50, ResNet-101, and ResNet-152 to empirically evaluate the performance of K-FAC as a preconditioner to SGD.

The CIFAR-10 dataset has 10 categories with a total of \num{50000} training images and \num{10000} validation images.
The ImageNet-1k dataset has \num{1000} categories with $\sim$1.3~M training images and \num{50000} validation images.

\subsection{Results}

\subsubsection{Correctness}
\begin{figure}[h]
\begin{center}
    \includegraphics[width=85mm]{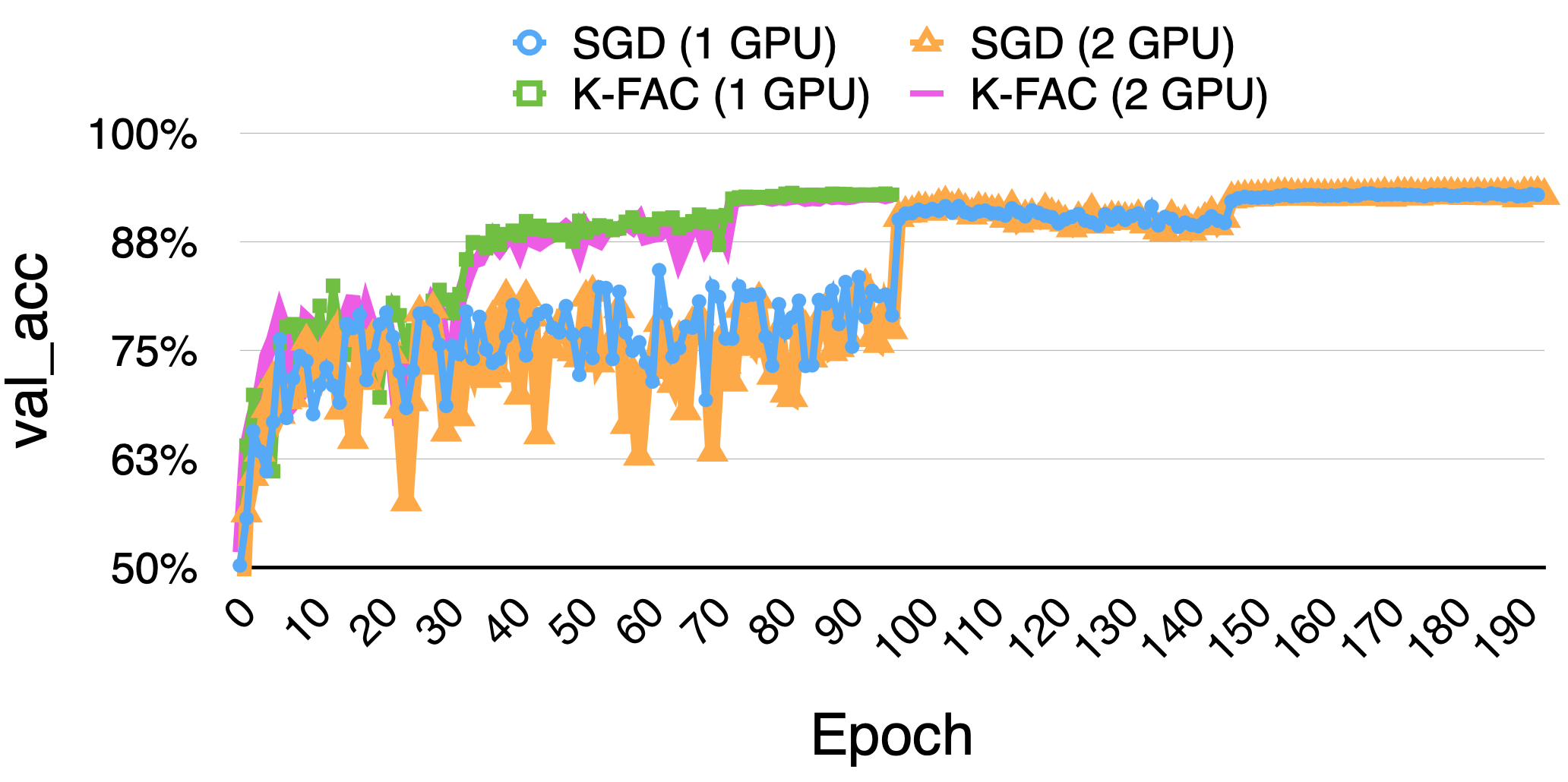}
    \caption{Validation accuracy comparison of ResNet-32 on CIFAR-10 with KFAC and SGD.
    \label{fig:cifar-res-kfac-sgd}}
\end{center}
\end{figure}

We first use CIFAR-10 and ResNet-34 to confirm the correctness of our K-FAC implementation. 
We adopt the baseline for acceptable accuracy to be 92.49\% \cite{he2016deep}.
We train with K-FAC for 100 epochs and SGD for 200 epochs.
For both optimization methods, we specify the learning rate as $N\times 0.1$ and the batch size as $N\times 128$ where $N$ is the number of GPUs. 
The learning rate is decreased by a factor of 10 at epochs 35, 75, 90 for K-FAC and 100, 150 for SGD, and a linear learning rate warmup is used for the first five epochs.
We maintain a constant K-FAC update frequency of 10 iterations.
Figure \ref{fig:cifar-res-kfac-sgd} shows the validation accuracy for SGD and K-FAC on one and two GPUs.
Table \ref{tb:cifar-sgd-vs-kfac} provides the final validation accuracies for SGD and K-FAC on 1, 2, 4, and 8 GPUs.
We find that our K-FAC implementation performs as well or better than SGD across a range of batch sizes, while converging in fewer iterations.

\begin{table}[h]
\begin{center}
    \caption{Validation accuracy comparison of ResNet-32 on CIFAR-10 with KFAC and SGD.}
    \begin{tabular}{ | c | c | c | c | c |}
    \hline
    GPUs  & 1       & 2       & 4       & 8     \\ \hline \hline
    SGD   & 92.76\% & 92.77\% & 92.58\% & 92.69\% \\ \hline
    K-FAC & 92.93\% & 92.76\% & 92.90\% & 92.92\% \\ \hline
    \end{tabular}
    \label{tb:cifar-sgd-vs-kfac}
\end{center}   
\end{table}

We train ResNet-50 on the ImageNet-1k dataset with K-FAC for 55 epochs and SGD for 90 epochs to confirm that K-FAC: (1) converges to the MLPerf 75.9\% validation accuracy baseline, (2) converges to an equal or higher validation accuracy than that achieved by SGD, and (3) converges in fewer iterations than SGD. 
In this experiment, we use the batch size of 32$\times$16=512 and the learning rate of 0.0125$\times$16 = 0.2, using a decay schedule at epoch 25, 35, 40, 45, and 50.
We specify damping value as 0.001 and evaluate K-FAC approximation every 10 iterations.
Labels are smoothed with a factor of 0.1.
For SGD, we set the momentum to 0.9.
For each test case, we use linear learning rate warmup for the first five epochs.

\begin{figure}[ht]
\begin{center}
    \includegraphics[width=85mm]{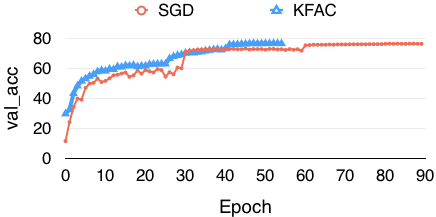}
    \caption{Validation accuracy comparison of ResNet-50 on ImageNet-1k with KFAC and SGD \edit{on 16 GPUs}. K-FAC converges to 76.4\% while SGD converges to 76.2\%. 
    \label{fig:res-kfac-sgd}}
\end{center}
\end{figure}

Figure~\ref{fig:res-kfac-sgd} shows the Top-1 validation accuracy curves of ResNet-50 on the ImageNet-1k dataset on 16 GPUs.
K-FAC as a preconditioner for SGD meets all three criteria by converging to 76.4\% validation accuracy, outperforming SGD with no preconditioning by 0.2\%, and consistently converging to the MLPerf baseline in under 55 epochs compared to the standard 90 epochs required by SGD.
In this case, K-FAC reaches 75.9\% in the 43rd epoch, compared to 76th epoch of SGD.

\subsubsection{Performance}

Reducing the frequency of computing the factors $A_{i-1}$ and $G_i$ and their eigen decompositions is key to achieving high-performance when training with K-FAC.
By reducing the K-FAC update frequency, we can avoid significant computation and communication at the cost of introducing stale information.
Understanding the tradeoff between staleness of information and improved time-to-solution is key when tuning the update frequency.

\begin{figure}[t]
\begin{center}
    \includegraphics[width=85mm]{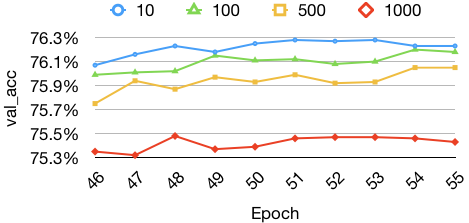}
    \caption{ResNet-50 validation accuracy of the last 10 epochs with K-FAC update frequency of \{10, 100, 500, 1000\} iterations. The MLPerf baseline is 75.9\%. 
    \label{fig:tuning-acc}}
\end{center}
\end{figure}

Table~\ref{tb:tuning} shows the Top-1 validation accuracy of ResNet-50, ResNet-101, and ResNet-152 trained with K-FAC for 55 epochs with a varying update frequency of \{100, 500, 1000\} iterations using 64 V100 GPUs.
Figure~\ref{fig:tuning-acc} shows the Top-1 validation accuracy in the last 10 epochs of each test case.
We find that ResNet-50 with all update frequencies except \num{1000} converge above the 75.9\% baseline. 
This consistent performance is key to achieving fast and scalable performance when training with K-FAC.
Observing that a larger interval than 500 iterations achieves marginal performance improvement,
we choose the K-FAC update interval as 500 iterations with 64 GPUs.
For the scalability experiments in \S\ref{sec:exp:scale}, we scale the interval according to the total batch size, so that the K-FAC update frequency is constant per epoch.
\edit{Specifically, we use {2000, 1000, 500, 250, 125}-iteration K-FAC update intervals for all ResNet scaling experiments on {16, 32, 64, 128, 256}-GPUs.
For ResNet-101 and ResNet-152, we observe a 0.2\% validation accuracy drop with K-FAC compared to SGD. 
Although there is no well-established baseline for ResNet-101 and ResNet-152, 
76.4\% and 76.6\% are documented in Keras Applications~\cite{keras-apps}, respectively.
Both our SGD and K-FAC results are significantly higher than these numbers.} 

\begin{table}[t]
\begin{center}
    \caption{\edit{ResNet-50, ResNet-101, and ResNet-152 validation accuracy vs.\ K-FAC update frequency with 64 GPUs.}}
    \begin{scriptsize}
    \begin{tabular}{ | c | c | c | c | c | c | }
    \cline{4-6}
    \multicolumn{3}{c|}{}& \multicolumn{3}{c|}{K-FAC Update Freq.} \\ \hline 
    \multicolumn{2}{|c|}{Model} & SGD & 100 & 500 & 1000 \\ \hline \hline
    \multirow{2}{*}{ResNet-50} & Val Accu & 76.2\% & 76.2\% & 76.1\% & 75.5\%  \\ \cline{2-6}
    & Train. Time (min) & 178 & 152 & 128 & 124 \\ \hline
    \multirow{2}{*}{ResNet-101} & Val Accu & 78.0\% & 77.7\% & 77.7\% & 77.3\%  \\ \cline{2-6}
    & Train. Time (min) & 244 & 227 & 197 & 195 \\ \hline
    \multirow{2}{*}{ResNet-152} & Val Accu & 78.2\% & 78.0\% & 78.0\% & 77.8\%  \\ \cline{2-6}
    & Train. Time (min) & 345 & 369 & 310 & 300 \\ \hline
    \end{tabular}
    \end{scriptsize}
    \label{tb:tuning}
\end{center}   
\end{table}

\subsubsection{Scalability}
\label{sec:exp:scale}
We test the scalability of our distributed K-FAC algorithm by measuring the time-to-solution on \{16, 32, 64, 128, 256\} GPUs. For each experiment, we measure the average time per epoch over 10 epochs and project the completed time span to 55 epoch for K-FAC and 90 epochs for SGD. 
Assuming we have $N$ GPUs, we specify the learning rate as $N\times0.0125$ the batch size of $N\times32$.
We set the damping value to 0.001 and the same learning rate decay schedule of {25, 35, 40, 45, 50} for K-FAC and {30, 40, 80} for SGD.
For both K-FAC and SGD, we use the linear learning rate warmup for the first five epochs.
We set the momentum of SGD to 0.9. 
We maintain the same hyper-parameters across all runs including the SGD hyper-parameters when SGD is used on its own and with K-FAC as a preconditioner.

To understand the performance benefits of our distribution and communication scheme for K-FAC, we compare two variants of our K-FAC preconditioner, \emph{K-FAC-lw} and \emph{K-FAC-opt}.
\emph{K-FAC-lw} refers to the K-FAC optimizer with just layer-wise distribution strategy and \emph{K-FAC-opt} represents the optimized distribution strategy.
Both variants use the K-FAC update procedure outlined in Equations \eqref{eq:v1}--\eqref{eq:kfac-eigen}; the only differences between the two are how work is distributed among workers and where communication occurs.
\emph{K-FAC-lw} uses the layer-wise distribution scheme of \cite{Osawa_2019_CVPR} where each worker computes the entire K-FAC update, i.e. the eigen decompositions of the factors and the preconditioned gradient, for a single layer and communicates the final preconditioned gradient for that layer to all other workers. 
\emph{K-FAC-opt} utilizes all of the optimizations introduced in this paper to reduce the frequency of communication by decoupling the eigen decomposition calculation from the preconditioning of the gradients.
\edit{We verify that all \emph{K-FAC-lw} and \emph{K-FAC-opt} experiments converge to validation accuracy of 76.2\%, 77.7\%, and 78.0\% for ResNet-50, ResNet-101, and ResNet-152, respectively.}

\begin{figure}[t]
\begin{center}
    \includegraphics[width=85mm]{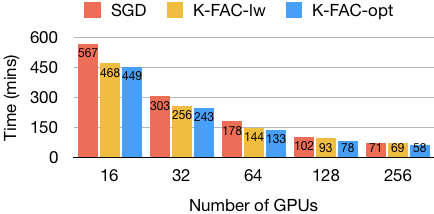}
    \caption{Time-to-solution comparison of ResNet-50 using K-FAC with SGD. 
    \label{fig:res50-kfac-sgd-scale}}
\end{center}
\end{figure}

Figure~\ref{fig:res50-kfac-sgd-scale} shows the time-to-solution comparison between K-FAC and SGD across scales. 
At all scales, K-FAC as a preconditioner to SGD converges to the MLPerf baseline in 55 epochs.
\emph{K-FAC-lw} outperforms SGD by 2.8-19.1\%, and \emph{K-FAC-opt} outperforms SGD by 17.7-25.2\%.
On 128 GPUs, the sustained scaling efficiency of \emph{K-FAC-opt} is 71.8\% which is a 9.4\% improvement over the 62.4\% efficiency of \emph{K-FAC-lw}.
It is also higher than the 68.6\% scaling efficiency of SGD.
The reduced communication frequency of \emph{K-FAC-opt} results in better scaling.
On 256 GPUs, the scaling efficiency of all three cases drop below 50\%.
However, \emph{K-FAC-lw} achieves 2.8\% improved performance than SGD whereas \emph{K-FAC-opt} yields an 18.3\% improvement at 256 GPUs.

\begin{figure}[t]
\begin{center}
    \includegraphics[width=85mm]{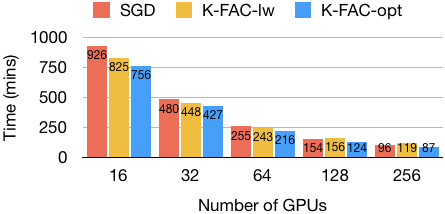}
    \caption{Time-to-solution comparison of ResNet-101 using K-FAC and SGD. 
    \label{fig:res101-kfac-sgd-scale}}
\end{center}
\end{figure}

\begin{figure}[t]
\begin{center}
    \includegraphics[width=85mm]{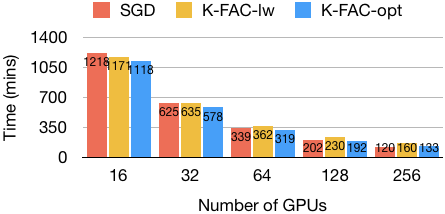}
    \caption{Time-to-solution comparison of ResNet-152 using K-FAC and SGD. 
    \label{fig:res152-kfac-sgd-scale}}
\end{center}
\end{figure}

\subsubsection{Limitations}
We compare the scaling of SGD and K-FAC across model sizes by training ResNet-50, ResNet-101, and ResNet-152 on the ImageNet-1k dataset.
Figure~\ref{fig:res101-kfac-sgd-scale} and Figure~\ref{fig:res152-kfac-sgd-scale} show the time-to-solution comparison between K-FAC and SGD across scales.
\emph{K-FAC-opt} outperforms SGD by 9.7-19.5\% on ResNet-101 at all scales and by 4.9-8.2\% on ResNet-152 up to 128 GPUs.
At 256 GPUs on ResNet-152, we find that \emph{K-FAC-opt} is 11.1\% slower than SGD.

\begin{table}[t]
\begin{center}
    \caption{Summary of \emph{K-FAC-opt} improvement over SGD}
    \begin{tabular}{ | c | c | c | c | c | c |}
    \hline
    Scale & 16 & 32 & 64 & 128 & 256 \\ \hline \hline
    ResNet-50 & 20.9\% & 19.7\% & 25.2\% & 23.5\% & 17.7\% \\ \hline
    ResNet-101 & 18.4\% & 11.1\% & 15.1\% & 19.5\% & 9.7\% \\ \hline
    ResNet-152 & 8.2\% & 7.6\% & 6.0\% & 4.9\% & -11.1\% \\ \hline
    \end{tabular}
    \label{tb:trend}
\end{center}   
\end{table}

Table~\ref{tb:trend} shows the improvement of \emph{K-FAC-opt} over SGD across the three models and scales.
\edit{We observe that \emph{K-FAC-opt}'s relative performance to SGD deteriorates with model complexity and scale.
To explain the observed scaling trend, we use the model in Figure~\ref{fig:streamline} in \S\ref{sec:back:sgd}.
The terms \emph{T}$_{i/o}$, \emph{T}$_f$, \emph{T}$_e$, \emph{T}$_x$, and \emph{T}$_u$ refer to the time cost of each of the five steps performed in each iteration: 1) I/O, 2) forward compute, 3) gradient evaluation, 4) gradient exchange, and 5) variable update.
To ease the discussion, we assume that these steps are performed in sequential order without any pipeline parallelism.}

\begin{figure}[H]
\begin{center}
    \includegraphics[width=85mm]{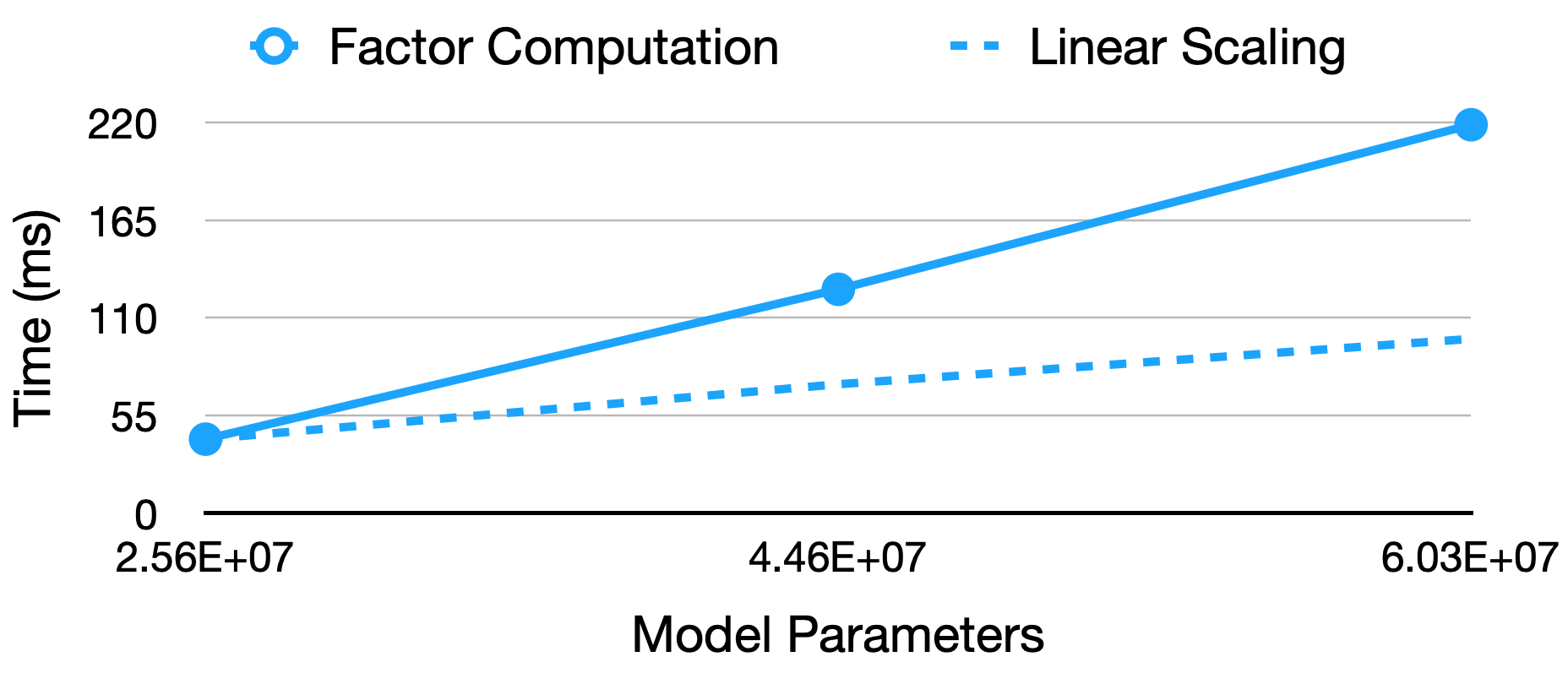}
    \caption{Factor computation time as model complexity increases. 
    \label{fig:fac_computation_time}}
\end{center}   
\end{figure}

\edit{The performance deterioration with increasing model complexity is attributed to the super-linear increase in \emph{T}$_e$:
as the model is more complex, the training time with SGD increases proportionally to the parameter count. 
With K-FAC, \emph{T}$_{i/o}$, \emph{T}$_f$, \emph{T}$_x$, and \emph{T}$_u$ consume identical time as SGD.
On the other hand, \emph{T}$_e$ in K-FAC includes two additional stages compared to SGD: factor computation and eigen decomposition.
We profile the factor computation cost per step on 16 V100 GPUs and observe a super-linear increase in time spent computing the factors $A_{i-1}$ and $G_i$ as model complexity increases, as shown in Figure \ref{fig:fac_computation_time}.
Unlike the eigen decomposition stage which can take advantage of the layer-wise distribution scheme, the factor computation time is constant as the GPU count is increased, as show in Table \ref{tb:kfac_profiling}, which leads to the super-linear increase in \emph{T}$_e$ and in turn to the relative performance deterioration of K-FAC relative to SGD.}

\begin{table}[t]
\begin{center}
    \caption{Time (ms) profile for the factor computation and eigen decomposition of a K-FAC update step across various model sizes and scales. \emph{T}$_{comm}$ is the communication time, and \emph{T}$_{comp}$ is the computation time.}
	\begin{tabular}{|c|c|c|c|c|c|}
	\cline{3-6}
	\multicolumn{1}{l}{}        &      & \multicolumn{2}{c|}{Factor}           & \multicolumn{2}{c|}{Eigen Decomposition} \\ \hline
	\multicolumn{1}{|c|}{Model} & GPUs & \emph{T}$_{comp}$ & \emph{T}$_{comm}$ & \emph{T}$_{comp}$ & \emph{T}$_{comm}$ \\ \hline \hline
	\multirow{3}{*}{ResNet-50}  & 16   & 36.83             & 155.79            & 2256.64           & 117.28          \\ \cline{2-6}
	                            & 32   & 43.30             & 171.57            & 1668.19           & 149.60          \\ \cline{2-6}
	                            & 64   & 44.90             & 154.63            & 1497.96           & 142.93          \\ \hline \hline
	\multirow{3}{*}{ResNet-101} & 16   & 125.23            & 224.15            & 3271.72           & 199.69          \\ \cline{2-6}
	                            & 32   & 126.14            & 267.08            & 2280.38           & 265.57          \\ \cline{2-6}
	                            & 64   & 126.95            & 239.33            & 2410.24           & 253.23          \\ \hline \hline
	\multirow{3}{*}{ResNet-152} & 16   & 218.36            & 276.83            & 4067.69           & 279.08          \\ \cline{2-6}
	                            & 32   & 219.00            & 313.17            & 2758.42           & 329.05          \\ \cline{2-6}
	                            & 64   & 219.12            & 312.52            & 2212.24           & 347.99          \\ \hline          
	\end{tabular}
    \label{tb:kfac_profiling}
\end{center}   
\end{table}

\begin{table}[t]
\begin{center}
    \caption{Minimum and maximum eigen decomposition worker speedup.}
    \begin{tabular}{| c | c | c | c | c | c | c |}
    \cline{2-7}
    \multicolumn{1}{l}{} & \multicolumn{2}{|c|}{ResNet-50} & \multicolumn{2}{c|}{ResNet-101} & \multicolumn{2}{c|}{ResNet-152} \\ \hline
    GPUs & min  & max  & min  & max  & min  & max  \\  \hline\hline
    16   & 1.00 & 1.00 & 1.00 & 1.00 & 1.00 & 1.00 \\  \hline
    32   & 1.34 & 2.88 & 1.41 & 3.33 & 1.51 & 2.03 \\  \hline
    64   & 1.55 & 6.61 & 1.26 & 6.18 & 1.85 & 8.27 \\  \hline
    \end{tabular}
    \label{tb:per_worker_speedup}
\end{center}   
\end{table}

\edit{To understand the performance deterioration at increasing scale, we examine the time spent computing the eigen decompositions.
The time required to complete the eigen decomposition stage is bounded by the slowest worker.
Ideally, as GPU count is doubled, the eigen decomposition time is halved because each GPU has approximately half the number of factors to decompose.
However, in ResNet models, the individual factors can vary widely in size, and thus while each worker may have roughly the same number of factors to decompose, there can be an imbalance in aggregate sizes.}

\edit{We can understand the factor size imbalance across workers by computing the total number of parameters assigned to each worker in ResNet-50.
On 16 GPUs, the minimum number of parameters assigned to a worker is $1.46\times 10^6$ and the maximum is $2.83 \times 10^7$.
In comparison, on 64 GPUs, the minimum parameter count is $1.64\times 10^4$ and the maximum is $2.26\times 10^7$.
It is clear that while some workers see a dramatic decrease in the number of parameters assigned, and thus a decrease in time spent decomposing the factors, some workers still have a similar amount of work, even though the number of workers was quadrupled.}

\edit{To further quantify this trend, we measure the time for each worker to eigen decompose its assigned factors for ResNet-50, 101, and 152 on 16, 32, and 64 GPUs. 
We record the times of the fastest and slowest worker in each iteration such that we can observe the relative decrease in eigen decomposition time for the fastest and slowest workers as the worker count is increased.
The minimum and maximum worker speedups are reported in Table \ref{tb:per_worker_speedup}.
Across all three models, the fastest workers saw a 6.18-8.27x speedup in time when moving from 16 to 64 GPUs; however, the slowest workers saw minimal speedups of 1.26-1.85x.
The imbalance in the work assigned to each GPU is exacerbated at scale as workers are left idle as they wait for the slower workers to finish the computation.}

\edit{Factors are distributed in a greedy, round-robin fashion which contributes to the imbalance in work assigned.
To resolve this scaling bottleneck, one direction is to implement a placement policy that uses factor size as a heuristic for the eigen decomposition time such that factors can be assigned to workers in a way that balances the time spent in this stage across workers.}

\section{Conclusion and Future Work}
\label{sec:conc}
We have presented a distributed K-FAC optimizer that is correct, efficient, and scalable.
The optimizer integrates techniques of layer-wise distribution, inverse-free second-order gradient evaluation, K-FAC approximation decoupling, and dynamic K-FAC update frequency.
We prototype the design using the widely adopted PyTorch and Horovod framework and release the source code under the permissive MIT license.
We empirically evaluate the correctness and scalability of the proposed optimizer using the classic ResNet model family with the CIFAR-10 and ImageNet-1k datasets.
Our results show that our optimizer converges to the 75.9\% MLPerf ResNet-50 baseline with 18--25\% less time than SGD.

In the future, we will explore alternative K-FAC approximation strategy that is more scalable than the current design. 
We will also design and evaluate solutions to avoid communications and reduce communication quantity to further enhance the scalability of K-FAC.

\section*{Acknowledgments}
\noindent
This work was supported in part by the U.S. Department of Energy, Office of Science, Advanced Scientific Computing Research, Contract DE-AC02-06CH11357, the Exascale Computing Project, Project Number 17-SC-20-SC, and by NSF OAC-1931354 and OAC-1818253.

%

\bibliographystyle{IEEEtran} 
\scriptsize
\balance
\bibliography{kfac}

\begin{thebibliography}{10}
\providecommand{\url}[1]{#1}
\csname url@samestyle\endcsname
\providecommand{\newblock}{\relax}
\providecommand{\bibinfo}[2]{#2}
\providecommand{\BIBentrySTDinterwordspacing}{\spaceskip=0pt\relax}
\providecommand{\BIBentryALTinterwordstretchfactor}{4}
\providecommand{\BIBentryALTinterwordspacing}{\spaceskip=\fontdimen2\font plus
\BIBentryALTinterwordstretchfactor\fontdimen3\font minus
  \fontdimen4\font\relax}
\providecommand{\BIBforeignlanguage}[2]{{%
\expandafter\ifx\csname l@#1\endcsname\relax
\typeout{** WARNING: IEEEtran.bst: No hyphenation pattern has been}%
\typeout{** loaded for the language `#1'. Using the pattern for}%
\typeout{** the default language instead.}%
\else
\language=\csname l@#1\endcsname
\fi
#2}}
\providecommand{\BIBdecl}{\relax}
\BIBdecl

\bibitem{you2018imagenet}
Y.~You, Z.~Zhang, C.-J. Hsieh, J.~Demmel, and K.~Keutzer, ``Image{N}et training
  in minutes,'' in \emph{47th International Conference on Parallel
  Processing}.\hskip 1em plus 0.5em minus 0.4em\relax ACM, 2018, p.~1.

\bibitem{Codreanu2017}
V.~Codreanu, D.~Podareanu, and V.~Saletore, ``Scale out for large minibatch
  {SGD}: Residual network training on {ImageNet-1K} with improved accuracy and
  reduced time to train,'' \emph{arXiv preprint arXiv:1711.04291}, 2017.

\bibitem{Akiba2017}
``Extremely large minibatch {SGD}: Training {ResNet-50 on ImageNet} in 15
  minutes.''

\bibitem{ying2018image}
C.~Ying, S.~Kumar, D.~Chen, T.~Wang, and Y.~Cheng, ``Image classification at
  supercomputer scale,'' \emph{arXiv preprint arXiv:1811.06992}, 2018.

\bibitem{mikami2018massively}
H.~Mikami, H.~Suganuma, Y.~Tanaka, Y.~Kageyama \emph{et~al.}, ``Massively
  distributed {SGD}: {ImageNet/ResNet-50} training in a flash,'' \emph{arXiv
  preprint arXiv:1811.05233}, 2018.

\bibitem{Osawa_2019_CVPR}
K.~Osawa, Y.~Tsuji, Y.~Ueno, A.~Naruse, R.~Yokota, and S.~Matsuoka,
  ``Large-scale distributed second-order optimization using kronecker-factored
  approximate curvature for deep convolutional neural networks,'' in \emph{IEEE
  Conference on Computer Vision and Pattern Recognition}, June 2019.

\bibitem{lee2019deepdrivemd}
H.~Lee, M.~Turilli, S.~Jha, D.~Bhowmik, H.~Ma, and A.~Ramanathan,
  ``{DeepDriveMD}: Deep-learning driven adaptive molecular simulations for
  protein folding,'' in \emph{IEEE/ACM 3rd Workshop on Deep Learning on
  Supercomputers}.\hskip 1em plus 0.5em minus 0.4em\relax IEEE, 2019, pp.
  12--19.

\bibitem{Carrasquilla2017}
J.~Carrasquilla and R.~G. Melko, ``Machine learning phases of matter,''
  \emph{Nature Physics}, 2017.

\bibitem{kates2019predicting}
J.~Kates-Harbeck, A.~Svyatkovskiy, and W.~Tang, ``Predicting disruptive
  instabilities in controlled fusion plasmas through deep learning,''
  \emph{Nature}, vol. 568, no. 7753, pp. 526--531, 2019.

\bibitem{mccandlish2018empirical}
S.~McCandlish, J.~Kaplan, D.~Amodei, and O.~D. Team, ``An empirical model of
  large-batch training,'' \emph{arXiv preprint arXiv:1812.06162}, 2018.

\bibitem{bottou2018optimization}
L.~Bottou, F.~E. Curtis, and J.~Nocedal, ``Optimization methods for large-scale
  machine learning,'' \emph{SIAM Review}, vol.~60, no.~2, pp. 223--311, 2018.

\bibitem{you2019large}
\BIBentryALTinterwordspacing
Y.~You, J.~Hseu, C.~Ying, J.~Demmel, K.~Keutzer, and C.-J. Hsieh, ``Large-batch
  training for {LSTM} and beyond,'' in \emph{Proceedings of the International
  Conference for High Performance Computing, Networking, Storage and Analysis},
  ser. SC ’19.\hskip 1em plus 0.5em minus 0.4em\relax New York, NY, USA:
  Association for Computing Machinery, 2019. [Online]. Available:
  \url{https://doi.org/10.1145/3295500.3356137}
\BIBentrySTDinterwordspacing

\bibitem{he2016deep}
K.~He, X.~Zhang, S.~Ren, and J.~Sun, ``Deep residual learning for image
  recognition,'' in \emph{IEEE Conference on Computer Vision and Pattern
  Recognition}, 2016, pp. 770--778.

\bibitem{devlin2018bert}
J.~Devlin, M.-W. Chang, K.~Lee, and K.~Toutanova, ``Bert: Pre-training of deep
  bidirectional transformers for language understanding,'' \emph{arXiv preprint
  arXiv:1810.04805}, 2018.

\bibitem{martens2015optimizing}
J.~Martens and R.~Grosse, ``Optimizing neural networks with kronecker-factored
  approximate curvature,'' in \emph{International conference on machine
  learning}, 2015, pp. 2408--2417.

\bibitem{ma2019inefficiency}
L.~Ma, G.~Montague, J.~Ye, Z.~Yao, A.~Gholami, K.~Keutzer, and M.~W. Mahoney,
  ``Inefficiency of k-fac for large batch size training,'' 2019.

\bibitem{mlperf}
``{MLPerf},'' \url{https://www.mlperf.org/}.

\bibitem{liu1989limited}
D.~C. Liu and J.~Nocedal, ``On the limited memory bfgs method for large scale
  optimization,'' \emph{Mathematical programming}, vol.~45, no. 1-3, pp.
  503--528, 1989.

\bibitem{NEURIPS2019_9015}
\BIBentryALTinterwordspacing
A.~Paszke, S.~Gross, F.~Massa, A.~Lerer, J.~Bradbury, G.~Chanan, T.~Killeen,
  Z.~Lin, N.~Gimelshein, L.~Antiga, A.~Desmaison, A.~Kopf, E.~Yang, Z.~DeVito,
  M.~Raison, A.~Tejani, S.~Chilamkurthy, B.~Steiner, L.~Fang, J.~Bai, and
  S.~Chintala, ``Pytorch: An imperative style, high-performance deep learning
  library,'' in \emph{Advances in Neural Information Processing Systems
  32}.\hskip 1em plus 0.5em minus 0.4em\relax Curran Associates, Inc., 2019,
  pp. 8024--8035. [Online]. Available:
  \url{http://papers.neurips.cc/paper/9015-pytorch-an-imperative-style-high-performance-deep-learning-library.pdf}
\BIBentrySTDinterwordspacing

\bibitem{sergeev2018horovod}
A.~Sergeev and M.~D. Balso, ``Horovod: Fast and easy distributed deep learning
  in {TensorFlow},'' \emph{arXiv preprint arXiv:1802.05799}, 2018.

\bibitem{Deng2009}
J.~Deng, W.~Dong, R.~Socher, L.-J. Li, K.~Li, and L.~Fei-Fei, ``Image{N}et: A
  large-scale hierarchical image database,'' in \emph{IEEE Conference on
  Computer Vision and Pattern Recognition}, 2009, pp. 248--255.

\bibitem{MLSL}
{Intel}, ``Intel(r) machine learning scaling library,'' 2019,
  \url{https://github.com/intel/MLSL}.

\bibitem{Abadi2016}
M.~Abadi, P.~Barham, J.~Chen, Z.~Chen, A.~Davis, J.~Dean, M.~Devin,
  S.~Ghemawat, G.~Irving, M.~Isard \emph{et~al.}, ``Tensor{F}low: A system for
  large-scale machine learning,'' in \emph{12th USENIX Symposium on Operating
  Systems Design and Implementation}, 2016.

\bibitem{incubator2017gloo}
``Gloo: Collective communications library with various primitives for
  multi-machine training,'' \url{https://github.com/facebookincubator/gloo}.

\bibitem{Thakur2005}
R.~Thakur, R.~Rabenseifner, and W.~Gropp, ``Optimization of collective
  communication operations in {MPICH},'' \emph{International Journal of High
  Performance Computing Applications}, vol.~19, no.~1, pp. 49--66, 2005.

\bibitem{ginsburg2018large}
B.~Ginsburg, I.~Gitman, and Y.~You, ``Large batch training of convolutional
  networks with layer-wise adaptive rate scaling,'' 2018.

\bibitem{recht2011hogwild}
B.~Recht, C.~Re, S.~Wright, and F.~Niu, ``Hogwild: A lock-free approach to
  parallelizing stochastic gradient descent,'' in \emph{Advances in Neural
  Information Processing Systems}, 2011, pp. 693--701.

\bibitem{zhang2015deep}
S.~Zhang, A.~E. Choromanska, and Y.~LeCun, ``Deep learning with elastic
  averaging {SGD},'' in \emph{Advances in Neural Information Processing
  Systems}, 2015, pp. 685--693.

\bibitem{gossiping-jin2016scale}
P.~H. Jin, Q.~Yuan, F.~Iandola, and K.~Keutzer, ``How to scale distributed deep
  learning?'' \emph{arXiv preprint arXiv:1611.04581}, 2016.

\bibitem{momentum-mitliagkas2016asynchrony}
I.~Mitliagkas, C.~Zhang, S.~Hadjis, and C.~R{\'e}, ``Asynchrony begets
  momentum, with an application to deep learning,'' in \emph{54th Annual
  Allerton Conference on Communication, Control, and Computing}.\hskip 1em plus
  0.5em minus 0.4em\relax IEEE, 2016, pp. 997--1004.

\bibitem{li2014scaling}
M.~Li, D.~G. Andersen, J.~W. Park, A.~J. Smola, A.~Ahmed, V.~Josifovski,
  J.~Long, E.~J. Shekita, and B.-Y. Su, ``Scaling distributed machine learning
  with the parameter server,'' in \emph{11th USENIX Symposium on Operating
  Systems Design and Implementation}, 2014, pp. 583--598.

\bibitem{alistarh2018convergence}
D.~Alistarh, C.~De~Sa, and N.~Konstantinov, ``The convergence of stochastic
  gradient descent in asynchronous shared memory,'' in \emph{ACM Symposium on
  Principles of Distributed Computing}.\hskip 1em plus 0.5em minus 0.4em\relax
  ACM, 2018, pp. 169--178.

\bibitem{martens2016conv}
R.~{Grosse} and J.~{Martens}, ``{A Kronecker-factored approximate Fisher matrix
  for convolution layers},'' \emph{arXiv e-prints}, p. arXiv:1602.01407, Feb.
  2016.

\bibitem{Chen2015}
T.~Chen, M.~Li, Y.~Li, M.~Lin, N.~Wang, M.~Wang, T.~Xiao, B.~Xu, C.~Zhang, and
  Z.~Zhang, ``{MXNet}: A flexible and efficient machine learning library for
  heterogeneous distributed systems,'' \emph{arXiv preprint arXiv:1512.01274},
  2015.

\bibitem{patarasuk2009bandwidth}
P.~Patarasuk and X.~Yuan, ``Bandwidth optimal all-reduce algorithms for
  clusters of workstations,'' \emph{Journal of Parallel and Distributed
  Computing}, vol.~69, no.~2, pp. 117--124, 2009.

\bibitem{Ba2017Distributed}
J.~Ba, R.~B. Grosse, and J.~Martens, ``Distributed second-order optimization
  using kronecker-factored approximations,'' in \emph{ICLR}, 2017.

\bibitem{martens2018kronecker}
J.~Martens, J.~Ba, and M.~Johnson, ``Kronecker-factored curvature
  approximations for recurrent neural networks,'' 2018.

\bibitem{wang2019eigen}
\BIBentryALTinterwordspacing
C.~Wang, R.~Grosse, S.~Fidler, and G.~Zhang, ``{E}igen{D}amage: Structured
  pruning in the {K}ronecker-factored eigenbasis,'' in \emph{Proceedings of the
  36th International Conference on Machine Learning}, vol.~97.\hskip 1em plus
  0.5em minus 0.4em\relax PMLR, 2019, pp. 6566--6575. [Online]. Available:
  \url{http://proceedings.mlr.press/v97/wang19g.html}
\BIBentrySTDinterwordspacing

\bibitem{george2018ekfac}
T.~George, C.~Laurent, X.~Bouthillier, N.~Ballas, and P.~Vincent, ``Fast
  approximate natural gradient descent in a kronecker-factored eigenbasis,'' in
  \emph{Proceedings of the 32nd International Conference on Neural Information
  Processing Systems}, ser. NIPS’18.\hskip 1em plus 0.5em minus 0.4em\relax
  Red Hook, NY, USA: Curran Associates Inc., 2018, p. 9573–9583.

\bibitem{KFACTF}
\BIBentryALTinterwordspacing
J.~Martens, ``Kfac-tensorflow,'' 2019. [Online]. Available:
  \url{https://github.com/tensorflow/kfac}
\BIBentrySTDinterwordspacing

\bibitem{krizhevsky2009learning}
A.~Krizhevsky, G.~Hinton \emph{et~al.}, ``Learning multiple layers of features
  from tiny images,'' University of Toronto, Technical Report TR-2009, 2009.

\bibitem{keras-apps}
``Keras applications,'' \url{https://keras.io/api/applications/}.

\end{thebibliography}


\end{document}